\documentclass[pdflatex,sn-mathphys-num]{sn-jnl}

\PassOptionsToPackage{round}{natbib}

\usepackage{hyperref}
\usepackage[dvipsnames]{xcolor}
\usepackage[utf8]{inputenc} 
\usepackage[T1]{fontenc}    
\usepackage{booktabs}       
\usepackage{amsmath}
\usepackage{amsfonts}      
\usepackage{nicefrac}       
\usepackage{microtype}     
\usepackage{multirow}
\usepackage{tabularx}     
\usepackage{multirow}    
\usepackage[table]{xcolor}  
\usepackage{array}       
\usepackage{graphicx}
\usepackage{booktabs}
\usepackage{geometry}
\usepackage{caption}
\usepackage{subcaption}
\usepackage{natbib}
\usepackage{amssymb} 
\usepackage{pifont}  
\usepackage{array}    
\usepackage{makecell} 
\usepackage{adjustbox}
\usepackage{rotating}
\usepackage{cleveref}
\usepackage{tablefootnote}
\usepackage{siunitx}
\usepackage{ifthen}
\usepackage{endnotes}
\usepackage{titlesec}

\title{Ensemble Learning of Machine Learning Force Fields}

\author[1]{\fnm{Bangchen} \sur{Yin}}\email{yinbc24@mails.tsinghua.edu.cn}
\equalcont{These authors contributed equally to this work.}
\author[1]{\fnm{Yue} \sur{Yin}}\email{yiny@mail.tsinghua.edu.cn}
\equalcont{These authors contributed equally to this work.}
\author[1]{\fnm{Yuda W.} \sur{Tang}}\email{twyd21@mails.tsinghua.edu.cn}
\author*[1]{\fnm{Hai} \sur{Xiao}}\email{haixiao@tsinghua.edu.cn}

\affil[1]{Department of Chemistry, Tsinghua University, Beijing 100084, China}

\begin{document}

\maketitle

\begin{abstract}

    Machine learning force fields (MLFFs) are a promising approach to balance the accuracy of quantum mechanics with the efficiency of classical potentials, yet selecting an optimal model amid increasingly diverse architectures that delivers reliable force predictions and stable simulations remains a core pratical challenge. Here we introduce EL-MLFFs, an ensemble learning framework that uses a stacking methodology to integrate predictions from diverse base MLFFs. Our approach constructs a graph representation where a graph neural network (GNN) acts as a meta-model to refine the initial force predictions. We present two meta-model architectures: a computationally efficient direct fitting model and a physically-principled conservative model that ensures energy conservation. The framework is evaluated on a diverse range of systems, including single molecules (methane), surface chemistry (methanol/Cu(100)), molecular dynamics benchmarks (MD17), and the MatPES materials dataset. Results show that EL-MLFFs improves predictive accuracy across these domains. For molecular systems, it reduces force errors and improves the simulation stability compared to base models. For materials, the method yields lower formation energy errors on the WBM test set. The EL-MLFFs framework offers a systematic approach to address challenges of model selection and the accuracy-stability trade-off in molecular and materials simulations.

\end{abstract}

\section{Introduction}

The potential energy surface (PES) is the cornerstone of molecular dynamics. For decades, force fields (FFs) have been indispensable for constructing approximate PESs, enabling simulations of complex systems at scales inaccessible to first-principles methods. From pioneering simulations of protein dynamics~\cite{McCammon1977} to modeling entire viral proteomes~\cite{Amaro2021SARS}, FFs have become a central tool in computational biology. This impact extends into materials science, where they have accelerated the discovery of novel materials like metal-organic frameworks (MOFs)~\cite{Wilmer2012NatChem}.

Despite their success, traditional FFs, such as AMBER~\cite{Cornell1995ff94} and CHARMM~\cite{MacKerell1998CHARMM22}, are limited by their fixed functional forms, which struggle to model polarization and cannot describe chemical reactions. While advancements like reactive FFs~\cite{vanDuin2001ReaxFF} have broadened their scope, a fundamental tension between accuracy and efficiency persists. Quantum mechanical methods, the benchmark for accuracy~\cite{Bartlett2007RMP}, remain too computationally expensive for large-scale simulations, creating a critical gap in the modeling landscape.

Machine learning force fields (MLFFs) have emerged as a powerful paradigm to bridge this gap, pioneered by the work of Behler and Parrinello~\cite{Behler2007PRL}. By learning from quantum mechanical data, MLFFs can represent high-dimensional potential energy surfaces (PESs) with near first-principles accuracy while maintaining computational tractability~\cite{Deringer2021AdvMat}. The field has evolved rapidly, from kernel methods like Gaussian Approximation Potentials (GAP)~\cite{Bartok2010GAP} to sophisticated graph-based~\cite{Schutt2018SchNet} and equivariant architectures~\cite{Batzner2022NequIP}, which explicitly incorporate physical symmetries. In recent years, the rapid development of pretrained models~\cite{yin2025alphanet,park20247net, yang2024mattersim, fu2025esen, batatia2022mace, dpa3, lysogorskiy2025grace} has revolutionized atomistic simulations by offering unprecedented fidelity.

However, the rapid proliferation of MLFFs has introduced new challenges. While energy predictions have become highly accurate, the precise prediction of atomic forces that are essential for dynamics remains a significant hurdle~\cite{Chmiela2017sGDML}. Furthermore, the sheer diversity of available models creates a "paradox of choice," making it difficult for researchers to select the optimal FF for their specific system. This situation highlights the need for a strategy to systematically harness the collective strengths of the existing MLFF ecosystem.

The fundamental strength of ensemble learning lies in its proven capacity to reduce overfitting and improve generalization by combining multiple diverse models~\cite{WOLPERT1992241}. This inherent robustness naturally extends to its application in uncertainty quantification, offering a powerful framework to assess model confidence. While initial studies have employed simple averaging or data-splitting techniques~\cite{uncertainty, uncertainty1, uncertainty2, uncertainty3}, these methods may not fully exploit the rich, complementary information captured by structurally diverse models. A more sophisticated approach is needed to learn the complex correlations between different model predictions and synthesize them into a more robust and accurate result.

Here we introduce the Stacking method to MLFFs as a systematic framework for investigating the performance ceiling of contemporary models. Our work explores how the synergistic integration of predictions from diverse base models can reveal an upper bound on force prediction accuracy and simulation stability, which is a limit not necessarily achievable by any single constituent model. Through the comparative study of a direct-fitting approach and a physically-principled conservative force field, we assess this collective potential and suggest that such an ensemble approach is a meaningful exploration, offering crucial insights that can guide the development of MLFFs by defining a clear and ambitious target for their performance.

\begin{figure}[ht]
   \centering
   \includegraphics[width=\textwidth]{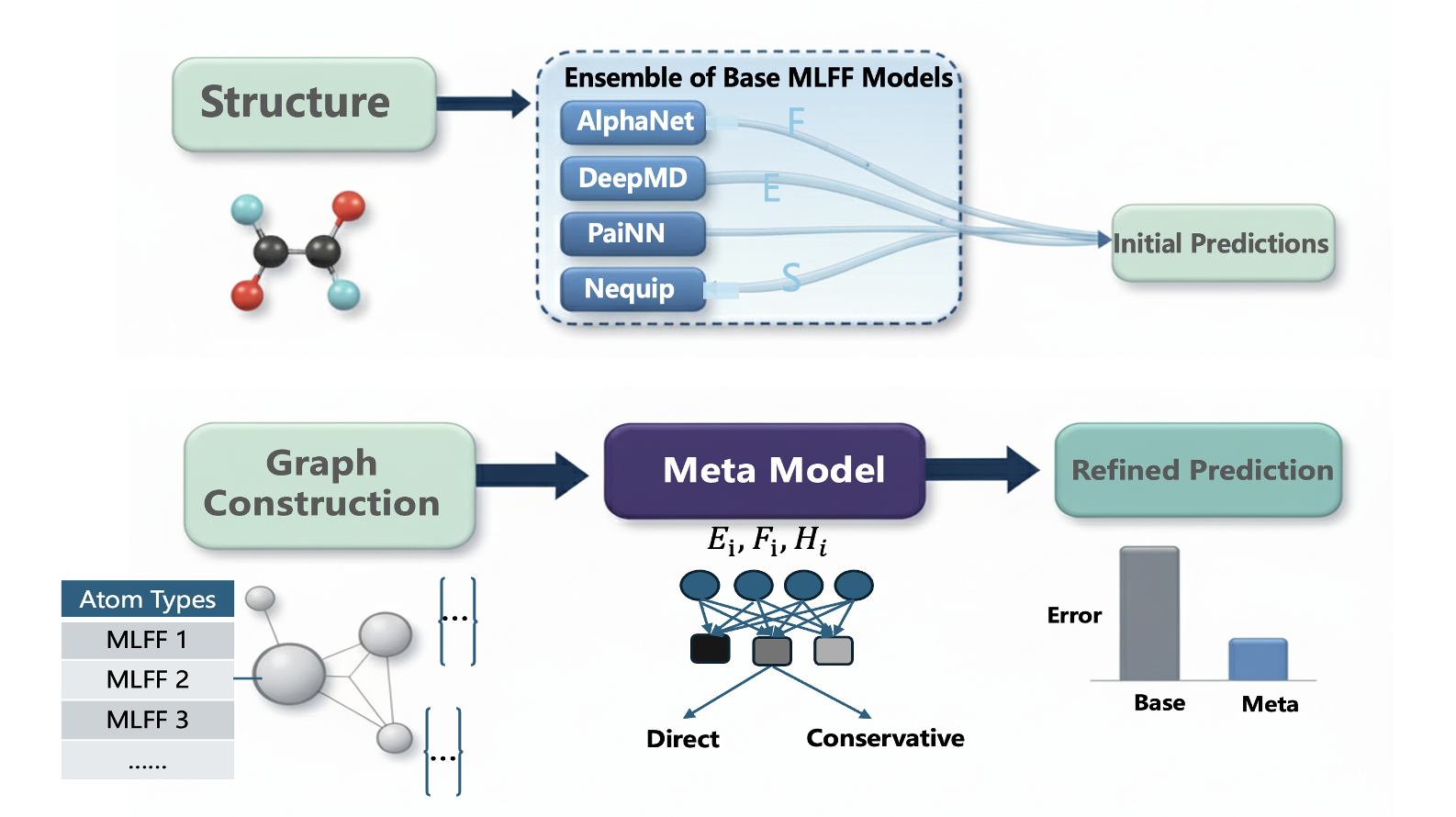}
   \caption{An overview of our ensemble learning architecture.}
   \label{fig:overall}
\end{figure}
Our work introduces a novel stacking-inspired ensemble framework designed to enhance the accuracy and reliability of MLFFs. Unlike traditional stacking, which treats base model predictions as a simple feature vector, our approach leverages the inherent graphical structure of molecules. As illustrated in Figure~\ref{fig:overall}, we first construct a graph for each molecule where the initial feature for each atom (node) is an aggregation of the forces predicted by a diverse set of pre-trained base models, supplemented with chemical information. A graph neural network (GNN) then serves as a meta-model, learning to refine these aggregated predictions to produce a final, more accurate force. We explore two distinct architectures for this GNN meta-model: a computationally efficient \textbf{direct-fitting} model (Ensemble-direct) based on a Graph Attention Network (GAT), and a physically-principled \textbf{conservative} model (Ensemble-conserv) that guarantees energy conservation by adapting an equivariant architecture.

\section{Results}
\subsection{Ensemble Framework Significantly Reduces Prediction Errors}

\begin{figure}[ht!]
    \centering
    \includegraphics[width=0.9\textwidth]{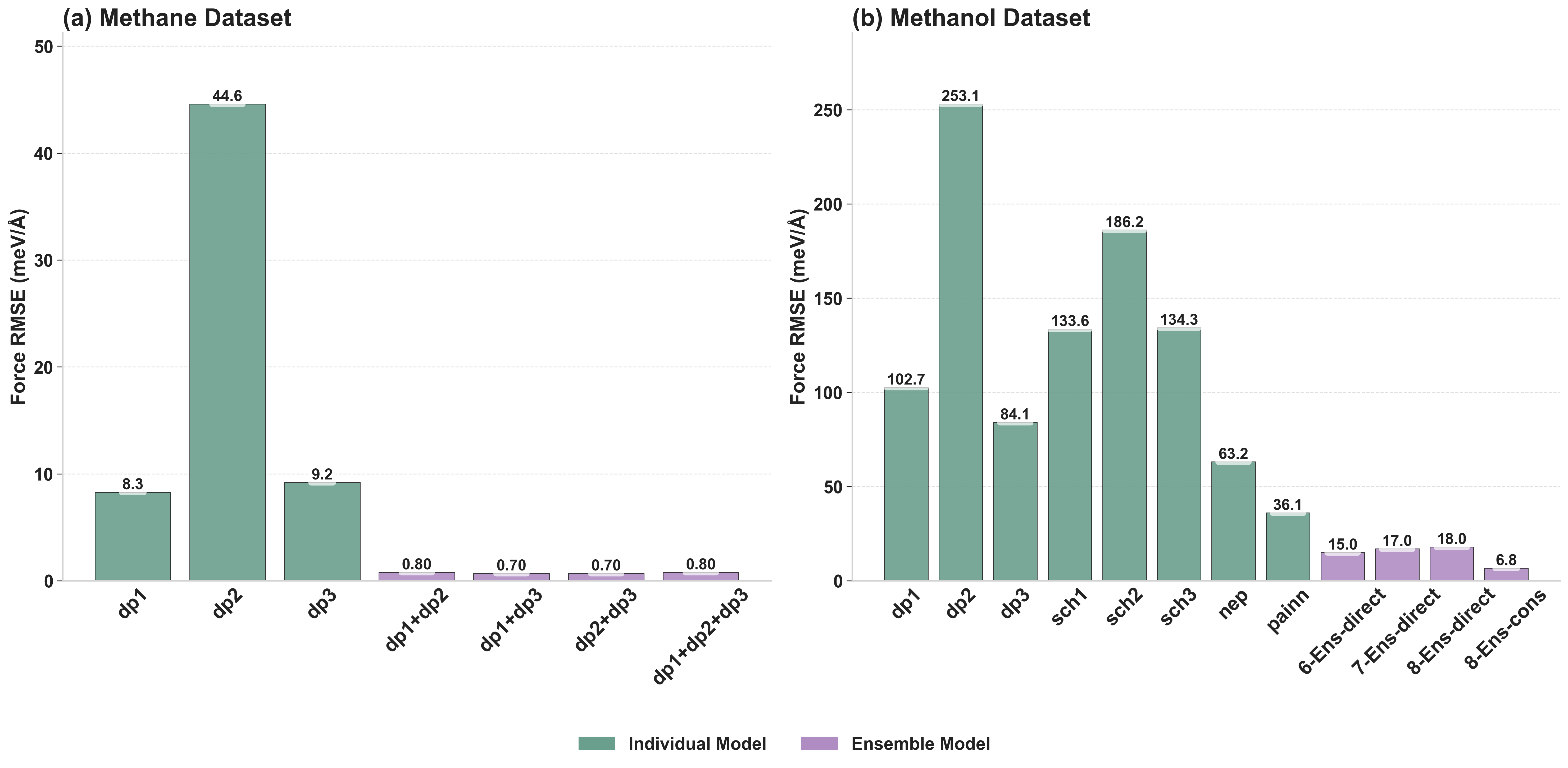}
    \caption{Performance comparison of individual (base) models and ensemble models on the methane and methanol datasets. \textbf{(a)} For the Methane dataset, a linear scale clearly illustrates the order-of-magnitude reduction in RMSE achieved by the ensemble method. \textbf{(b)} For the more complex Methanol dataset, a logarithmic scale is used, highlighting that ensemble errors are thousands of times lower than those of the individual models.}
    \label{fig:performance_comparison}
\end{figure}

To demonstrate the efficacy and scalability of our ensemble framework, we evaluated its performance on two distinct systems: a simple methane molecule and a more complex methanol-on-copper surface. The results, summarized in Figure~\ref{fig:performance_comparison}, reveal a dramatic improvement in force prediction accuracy that spans several orders of magnitude.

As shown in Figure~\ref{fig:performance_comparison}a for the methane dataset, our ensemble method reduces the force RMSE by an order of magnitude, from a range of 8.3-44.6 meV/Å for individual models down to a consistent range of 0.7-0.8 meV/Å. This performance leap is even more pronounced for the challenging methanol system (Figure~\ref{fig:performance_comparison}b), where a logarithmic scale is necessary to visualize the vast disparity in accuracy. While the base models exhibit RMSEs from 36.1 to 253.1 meV/Å, our ensemble approach achieves exceptionally low RMSEs of 15-18 meV/Å. This represents a 2-fold to 17-fold reduction in RMSE compared to the individual base models. Furthermore, RMSE can be reduced to just 6.8 meV/Å by employing a more expressive conservative model.

\begin{figure}[ht!]
    \centering
    \includegraphics[width=\textwidth]{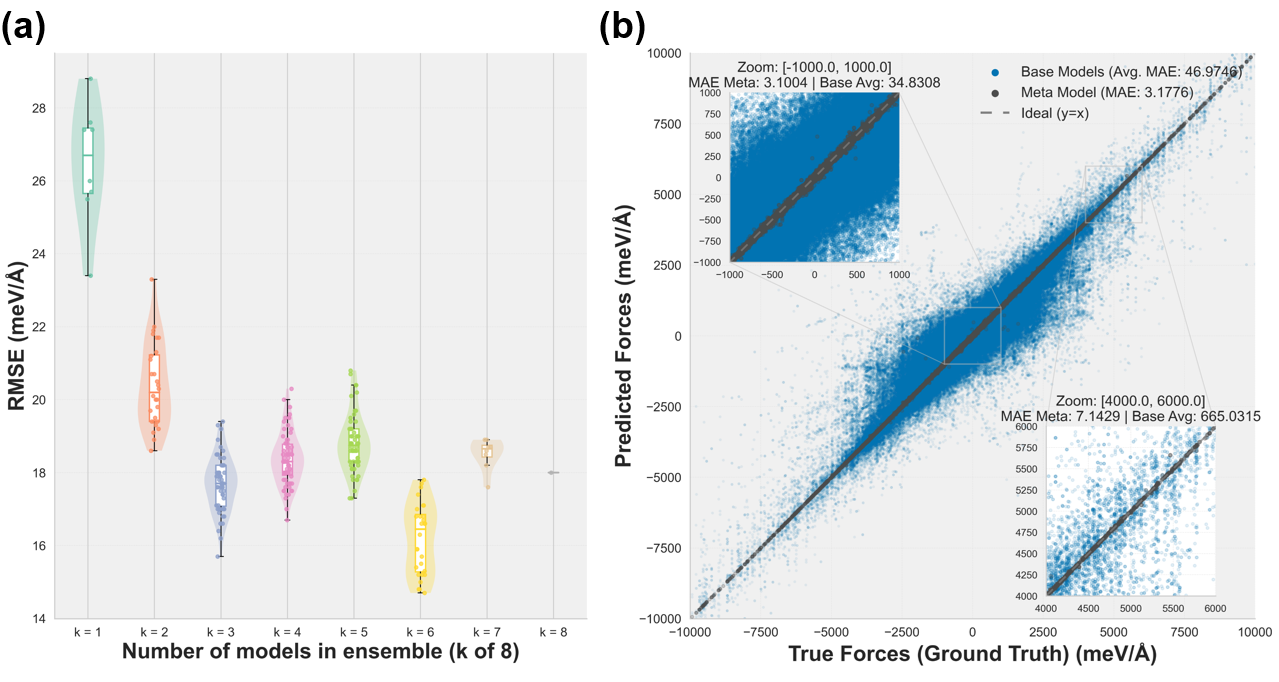}
    \caption{(a) Raincloud plot of RMSE values for all possible ensemble combinations on the methanol dataset, grouped by ensemble size $k$ (out of 8 base models). (b) Parity plot comparing predicted forces from the 8-model conservative ensemble to reference forces for the methane test set.}
    \label{fig:ensemble_combined}
\end{figure}

To further examine the synergistic behavior within the ensemble for the methanol system, we evaluated the performance fluctuations across all possible subsets of the eight base models. The resulting trends are summarized in Figure~\ref{fig:ensemble_combined}, which brings together the ensemble-scale error statistics and the force-prediction performance into a single coherent visualization. In the Figure~\ref{fig:ensemble_combined} (a), a raincloud plot shows the distribution of RMSE values (in meV Å$^{-1}$) for each ensemble size. This representation captures both the overall structure of the error landscape and the variability across different model combinations, enabling a detailed assessment of how predictive accuracy evolves as additional models are incorporated. We observe a marked reduction in both median error and spread with increasing ensemble size, reflecting the capacity of larger ensembles to suppress noise and enhance robustness. However, the improvements begin to plateau beyond a certain threshold—k = 6 in our case—indicating diminishing returns and the onset of redundancy when additional models are incorporated. This trend underscores that the effectiveness of our approach arises not from numerical aggregation alone, but from identifying an optimal degree of complementarity that balances predictive efficiency with computational practicality.

The Figure~\ref{fig:ensemble_combined} (b) further assesses the optimized eight-model ensemble using the methane test set. The parity plot compares predicted and reference forces over the full [$-$10, 10] eV Å$^{-1}$ range, with inset magnifications highlighting the [$-$1, 1] eV Å$^{-1}$ and [4, 6] eV Å$^{-1}$ regions. Across these regimes, the limitations of individual base models become evident: the mean absolute error increases sharply from 34.8 meV Å$^{-1}$ in the low-force region to 665 meV Å$^{-1}$ in the high-force region, representing nearly a twenty-fold degradation and revealing the poor reliability of single models for strongly non-equilibrium configurations. In contrast, the ensemble consistently maintains errors on the order of $1$ meV Å$^{-1}$ across both low- and high-force regions, demonstrating substantial gains in accuracy together with enhanced stability and generalization. Collectively, these results reveal that a carefully constructed ensemble can deliver predictive performance that surpasses the capability of any individual constituent, particularly for challenging non-equilibrium states where accurate force prediction is most critical.

\subsection{Ensemble Models Overcome the Accuracy-Stability Trade-off}

\subsubsection{Results of our method}

\begin{table}[htbp]
\centering
\caption{Performance comparison of different models across four molecular backgrounds. The metrics used are MAE of forces (meV/\AA), Equation~\ref{tab:stability} for determining the percentage of stability (\%), and Equation~\ref{eq:hr} for MAE of $h(r)$. The arrows indicate whether lower ($\downarrow$) or higher ($\uparrow$) values are better. The best performance in each category is highlighted in bold.}
\small
\setlength{\tabcolsep}{4pt} 
\begin{tabularx}{\textwidth}{@{} l l *{7}{>{\centering\arraybackslash}X} @{}}
\toprule
\rowcolor{gray!20}
\textbf{Molecule} & \textbf{Metric} & \textbf{DeepMD} & \textbf{SchNet} & \textbf{ForceNet} & \textbf{DimeNet} & \textbf{NequIP} & \textbf{Ensemble-direct} & \textbf{Ensemble-conserv} \\
\midrule
\multirow{3}{*}{\textbf{Aspirin}}
& forces\textsuperscript{$\downarrow$} & 35.1 & 44.7 & 32.1 & 24.0 & 15.2 & 8.9 & \textbf{8.5} \\
& stability\textsuperscript{$\uparrow$} & 67 & 1.8 & 12 & \textbf{100} & \textbf{100} & \textbf{100} & \textbf{100} \\
& \textit{h(r)}\textsuperscript{$\downarrow$} & 0.41 & 0.09 & \textbf{0.03} & 0.04 & 0.04 & 0.04 & 0.04 \\
\midrule
\multirow{3}{*}{\textbf{Ethanol}}
& forces\textsuperscript{$\downarrow$} & 30.0 & 35.7 & 39.9 & 29.8 & 18.1 & 6.8 & \textbf{6.5} \\
& stability\textsuperscript{$\uparrow$} & \textbf{100} & 88 & 43 & \textbf{100} & \textbf{100} & \textbf{100} & \textbf{100} \\
& \textit{h(r)}\textsuperscript{$\downarrow$} & 0.05 & 0.03 & \textbf{0.02} & 0.04 & 0.03 & \textbf{0.02} & 0.03 \\
\midrule
\multirow{3}{*}{\textbf{Naphthalene}}
& forces\textsuperscript{$\downarrow$} & 23.2 & 31.7 & 28.5 & 21.0 & 12.5 & 5.4 & \textbf{5.0} \\
& stability\textsuperscript{$\uparrow$} & 17 & 1.3 & \textbf{100} & 23 & \textbf{100} & \textbf{100} & \textbf{100} \\
& \textit{h(r)}\textsuperscript{$\downarrow$} & 0.06 & 0.23 & 0.04 & 0.08 & 0.04 & 0.04 & \textbf{0.03} \\
\midrule
\multirow{3}{*}{\textbf{Salicylic Acid}}
& forces\textsuperscript{$\downarrow$} & 28.8 & 36.8 & 29.5 & 23.9 & 16.3 & 8.2 & \textbf{8.0} \\
& stability\textsuperscript{$\uparrow$} & 74 & 23 & 48 & 52 & 95 & \textbf{100} & \textbf{100} \\
& \textit{h(r)}\textsuperscript{$\downarrow$} & 0.09 & 0.08 & 0.07 & \textbf{0.04} & 0.05 & 0.05 & 0.05 \\
\bottomrule
\end{tabularx}
\label{tab:model_comparison}
\end{table}

The results presented in Table~\ref{tab:model_comparison} clearly demonstrate that our proposed ensemble methods consistently outperform the individual base MLFFs across all four molecular systems. This superior performance is evident not only in numerical accuracy but also in the crucial metric of simulation stability.

A persistent challenge in the development of MLFFs is the trade-off between force prediction accuracy and long-term simulation stability. For instance, while DimeNet offers the best force predictions among the base models for Aspirin (24.0~meV/\AA), our conservative ensemble (Ensemble-conserv) reduces this error nearly threefold to just \textbf{8.5~meV/\AA}. This trend holds for all molecules, with our ensemble methods consistently achieving force MAEs below 10~meV/\AA, a significant improvement over the base models, whose errors typically range from 20 to 45~meV/\AA.

Perhaps most critically, our approach resolves instabilities observed in the base models. While individual models like SchNet and DeepMD show poor stability for molecules like Naphthalene (1.3\% and 17\%, respectively), both of our ensemble methods achieve \textbf{100\% stability} in nearly every case, matching or surpassing the best-performing individual model for each molecule. This robust stability, combined with high force accuracy, indicates that our method effectively mitigates the accuracy-stability dilemma by leveraging the collective strengths of its constituent models. Further evidence of this is provided by the MAE of \textit{h(r)}, where our ensembles consistently rank among the best performers, signifying a more accurate representation of the underlying physical particle distributions.

The primary advantage of our ensemble learning approach is the significant enhancement of simulation stability, in addition to improved predictive accuracy. By averaging predictions across multiple models, the framework smooths the potential energy surface, effectively mitigating outlier predictions and unphysical energy artifacts from any single model. This prevents the simulation failures often encountered with single-model force fields, enabling more robust and extended molecular dynamics trajectories. This dual benefit of accuracy and stability is achieved with a manageable increase in computational overhead, making it a practical solution for complex systems in drug discovery and materials science.

\subsection{Ensemble Approach Demonstrates Superior Scalability and Generalization}
To evaluate the scalability and generalization capabilities of our proposed methods, we utilized the large-scale MatPES PBE dataset, which contains 434,712 structures. We assessed the models' performance on the 1,000 strucrures from the WBM test set, a challenging task designed to measure out-of-domain accuracy. In these structural relaxations, our ensemble models predicted energies and forces, while the stress tensor was sourced from the CHGNet base model.

The results, summarized in Table~\ref{tab:matpes}, clearly demonstrate the effectiveness of our ensemble strategies. Among the base models, CHGNet exhibited the strongest performance, particularly with a leading forces MAE starting at 81 meV/\AA. However, both of our ensemble models consistently surpassed the performance of all individual base models across every metric. The direct ensemble method (Ensemble-direct) reduced the energy MAE to as low as 20 meV/atom.

Notably, our conservative ensemble model Ensemble-conserv achieved the most significant improvements in the critical metric of formation energy. It lowered the $E_f$ MAE to 0.071 eV/atom, a substantial reduction from the 0.080 eV/atom achieved by the best-performing base model, TensorNet. This demonstrates that our ensemble approach not only enhances predictive accuracy for fundamental properties like energy and forces but also improves its capability for crucial downstream tasks like predicting material stability. These findings validate that our method effectively scales to large datasets and improves out-of-domain generalization, yielding more robust and accurate force fields.

\begin{table}[ht]
\centering
\caption{Comparison of MAE on the MatPES and WBM datasets. The best performance in each category is highlighted in bold.}
\label{tab:matpes}
\begin{tabular}{l c c S[table-format=1.3]}
\toprule
\textbf{Model} & \textbf{ENERGY MAE} & \textbf{FORCES MAE} & {\textbf{$E_f$ MAE}} \\
 & \textbf{(meV/atom)} & \textbf{(meV/\AA)} & \textbf{(eV/atom)} \\
\midrule
M3GNet        & 40 / 45 / 45 & 155 / 177 / 181 & 0.110 \\
TensorNet     & 33 / 36 / 36 & 121 / 138 / 148 & 0.080 \\
CHGNet        & 27 / 32 / 31 & \textbf{81} / 124 / 136  & 0.081 \\
AlphaNet (Not in the ensemble)      & 22 / 26 /27 & 101 / 121 / 131  & 0.077 \\
Ensemble-direct & 23 / 26 / 27 & 112 / 117 / \textbf{123} & 0.074 \\
Ensemble-conserv & \textbf{20} / \textbf{24} / \textbf{26} & 110 / \textbf{115} / \textbf{123} & {\textbf{0.071}} \\
\bottomrule
\end{tabular}
\end{table}

\subsection{Computational Analysis Reveals a Practical Trade-off Between Efficiency and Physical Fidelity}
To demonstrate the practicality of our method, we must consider the computational cost and speed. Achieving the higher precision and stability offered by our method inherently involves the computational expense of the base MLFFs utilized. However, the crucial aspect to evaluate is the additional cost introduced by our meta-model. Specifically, we need to quantify the extra computational overhead and time required by our approach compared to the base models alone. Understanding this additional cost is essential for assessing the overall efficiency and feasibility of implementing our method in real-world scenarios.

\begin{table}[ht]
\centering
\caption{Comparison of computational cost by different models. Training time is reported in hours (h) and inference time in seconds (s) per structure. The best performance in each category is highlighted in bold.}
\label{tab:cost}
\small 
\setlength{\tabcolsep}{4pt}
\begin{tabular}{@{}lccccccc@{}}
\toprule
\textbf{Models} & \textbf{DeepMD} & \textbf{DimeNet} & \textbf{SchNet} & \textbf{ForceNet} & \textbf{NequIP} & \textbf{\makecell{Ensemble-\\direct}} & \textbf{\makecell{Ensemble-\\conserv}} \\
\midrule
\textbf{Train (h)} & 37.2 & 26.83 & 3.61 & 6.39 & 45.0 & \textbf{0.47} & 52.5 \\
\textbf{Inference (s)} & 0.62 & 0.91 & 0.53 & 0.31 & 1.10 & \textbf{0.037} & 1.35 \\
\bottomrule
\end{tabular}
\end{table}
 
Beyond predictive accuracy, the computational efficiency of our methods is a critical factor for their practical utility. As detailed in Table~\ref{tab:cost}, our ensemble approaches present distinct cost profiles. The Ensemble-direct model is remarkably efficient, with a training time of only 0.47 hours, an order of magnitude faster than training any of the base models from scratch. Its inference time is also exceptionally low at 0.037 seconds, making it ideal for large-scale screening or applications where speed is paramount.

This efficiency, however, presents a clear trade-off when strict physical principles must be upheld. The Ensemble-conserv model, by design, guarantees a conservative force field. This physically-rigorous constraint comes at a computational cost, making it the most demanding model in our study, both in terms of training time (52.5 hours) and inference speed (1.35 seconds). This increased overhead is a direct consequence of the more complex architecture and input requirements (such as Hessians) needed to enforce energy conservation via the chain rule.

Ultimately, this trade-off offers valuable flexibility. For tasks where rapid prediction is the primary goal, the Ensemble-direct method provides a powerful and highly efficient solution. On the other hand, for applications where physical fidelity is non-negotiable, such as in NVE simulations or phonon calculations, the Ensemble-conserv method provides the necessary guarantees, justifying its higher computational budget.

\section{Method}

\subsection{Stacking}

Stacked generalization, or \textit{stacking}, employs multiple base models and a meta-model to enhance predictive accuracy. The methodology unfolds across two primary stages:

\textbf{Stage 1: Base Model Predictions.} Given a dataset

\begin{equation}
D = \{(x_i, y_i)\}_{i=1}^N,
\end{equation}

where $x_i$ represents the feature vector and $y_i$ the corresponding target, a collection of $M$ base models $\{f_m\}_{m=1}^M$ is trained. Each base model $f_m: \mathbb{R}^d \rightarrow \mathbb{R}$ (for regression) or $f_m: \mathbb{R}^d \rightarrow \{c_1, c_2, \ldots, c_k\}$ (for classification) generates a prediction $\hat{y}_{i,m} = f_m(x_i)$ for each instance $x_i$.

\textbf{Stage 2: Meta-Model Integration.} The predictions from the base models are aggregated into a new feature vector for each instance,

\begin{equation}
\tilde{x}_i = [\hat{y}_{i,1}, \hat{y}_{i,2}, \ldots, \hat{y}_{i,M}],
\end{equation}

resulting in a transformed dataset $\tilde{D} = \{(\tilde{x}_i, y_i)\}_{i=1}^N$. A meta-model $g: \mathbb{R}^M \rightarrow \mathbb{R}$ (for regression) or $g: \mathbb{R}^M \rightarrow \{c_1, c_2, \ldots, c_k\}$ (for classification) is subsequently trained on $\tilde{D}$ to predict the target $y$ based on the meta-features $\tilde{x}$.

\subsection{Graph Construction}

In our framework, each molecular structure is represented as a graph $G_i = (V_i, E_i)$, where atoms are the nodes ($V_i$) and bonds are the edges ($E_i$). The initial features for each node are constructed by leveraging predictions from an ensemble of $M$ pre-trained MLFFs, denoted as $\{f_m\}_{m=1}^M$.

For a given molecular structure $i$, each base model $f_m$ provides two outputs: a scalar prediction for the total potential energy of the molecule, $E_i^m$, and a predicted force vector for each individual atom $j$ within that molecule, which we denote as $\hat{\mathbf{F}}_{i,j}^m$.

The initial feature representation for a specific node $j$ in the graph of molecule $i$, denoted as $\mathbf{h}_{i,j}$, is formulated by concatenating these predictions with the atom's chemical identity (e.g., its atomic number). This feature representation effectively combines \textbf{local information} (the forces predicted on that particular atom) with \textbf{global information} (the total energy of the entire structure):
\begin{equation}
\mathbf{h}_{i,j} = \text{Concat}\left( [\hat{\mathbf{F}}_{i,j}^1, \hat{\mathbf{F}}_{i,j}^2, \ldots, \hat{\mathbf{F}}_{i,j}^M], [E_i^1, E_i^2, \ldots, E_i^M], \text{atomic\_identity}_j \right)
\end{equation}
The predicted energies are typically passed through an embedding layer before concatenation to create a richer feature representation. This process yields a comprehensive set of initial node features that captures the collective knowledge of the entire MLFF ensemble for each atom in the context of its molecular environment.

\subsection{Model Architecture}

This section details the two primary architectural frameworks explored in our work. The first is a \textbf{direct-fitting} model, which leverages a GAN with residual connections and a jumping knowledge mechanism for end-to-end learning. The second framework introduces a method for constructing a \textbf{conservative force field}, which ensures physical consistency by applying the chain rule to integrate the energy, force, and Hessian predictions from multiple pre-trained equivariant base models. A key distinction lies in their input features: the direct-fitting model utilizes only the force predictions from the base models and atomic species information, whereas the conservative force field framework also incorporates geometric structural features.

\subsubsection{Directly Fitting}

For the direct fitting approach, we employ a GAT \cite{veličković2018graph}. To stabilize the training process and improve gradient flow, \textbf{residual connections} \cite{RESNET} are integrated into each GAT layer. The node representation update for the $l$-th layer is formulated as:
\begin{equation}
    \mathbf{H}^{(l+1)} = \mathbf{H}^{(l)} + \sigma(\text{GATConv}(\mathbf{H}^{(l)}, \mathbf{E}))
    \label{eq:gat_residual}
\end{equation}
where $\mathbf{H}^{(l)}$ is the node feature matrix at layer $l$, $\mathbf{E}$ contains the graph's adjacency information, and $\sigma$ denotes the SiLU activation function.

Following message passing, we utilize a \textbf{Jumping Knowledge (JK)} mechanism \cite{xu2018representation} to aggregate node representations from all layers. This allows the model to capture multi-scale structural information. Specifically, we concatenate the outputs from all $L$ layers to form the final node representation, $\mathbf{H}_{\text{final}}$:
\begin{equation}
    \mathbf{H}_{\text{final}} = \text{Concat}\left([\mathbf{H}^{(1)}, \mathbf{H}^{(2)}, \ldots, \mathbf{H}^{(L)}]\right)
    \label{eq:jk}
\end{equation}

\subsubsection{Conservative Force Field Implementation}
\label{sec: conserv}
A \textbf{conservative force field} is essential for applications such as molecular dynamics simulations in the NVE ensemble and for phonon calculations. A simple baseline approach is to linearly combine the predictions of multiple base models, for instance, by averaging their outputs. However, the accuracy of such a method is inherently limited by that of the most precise base model, as it fails to learn from the combined predictions.

To overcome this limitation, we propose a more sophisticated method that constructs a conservative force field by applying the \textbf{chain rule} to a learned potential energy function. Let $\theta$ be a learned function that maps the outputs of the base models (energy $E_i$, forces $\mathbf{F}_i$) and the atomic coordinates $\mathbf{x}$ to a total potential energy, $E_{\text{total}} = \theta(E_1, \mathbf{F}_1, \ldots, E_L, \mathbf{F}_L, \mathbf{x})$. The total force $\mathbf{F}_{\text{total}}$ is the negative gradient of this potential. Applying the multivariate chain rule yields:
\begin{equation}
    \mathbf{F}_{\text{total}} = -\nabla_{\mathbf{x}}E_{\text{total}} = -\frac{\partial \theta}{\partial \mathbf{x}} + \sum_{i=1}^{L} \left( \frac{\partial \theta}{\partial E_i} \mathbf{F}_i + \frac{\partial \theta}{\partial \mathbf{F}_i} \mathbf{H}_i \right)
    \label{eq:conservative_force}
\end{equation}
where $\mathbf{F}_i = -\nabla_{\mathbf{x}}E_i$ and $\mathbf{H}_i = -\nabla_{\mathbf{x}}\mathbf{F}_i$ are the force and the negative of the Hessian matrix predicted by the $i$-th base model, respectively. This formulation guarantees that the resulting force field is conservative, as it is derived from a single scalar potential $\theta$.

To assess the contribution of the force components as input features, we also perform an ablation study in the supplementary materials using a simplified variant of this model. In this setup, the meta-model $\theta$ is conditioned only on the energies of the base models, such that $E_{\text{total}} = \theta(E_1, \ldots, E_L, \mathbf{x})$. The resulting chain rule for the total force is simplified and no longer requires the Hessian matrices from the base models:
\begin{equation}
    \mathbf{F}_{\text{total}} = -\frac{\partial \theta}{\partial \mathbf{x}} + \sum_{i=1}^{L} \frac{\partial \theta}{\partial E_i} \mathbf{F}_i
    \label{eq:conservative_force_ablation}
\end{equation}
This comparison allows us to quantify the value of including explicit force information in the meta-model and highlights the trade-off between accuracy and the computational overhead associated with calculating Hessians.

The meta-model $\theta$, which combines the predictions from the base models, is itself designed to be equivariant. To this end, we implemented distinct architectures for this purpose, which built upon the spherical harmonics framework of \textbf{NequIP}.

\subsection{Settings}
\label{sec:setting}

To evaluate the effectiveness and versatility of our method, we benchmarked its performance on a diverse range of chemical systems. Our evaluation includes two custom datasets: a simple test case of single methane molecules and a more complex system of methanol molecules adsorbed on a three-layer Cu(4x4) (100) surface.

We further assess our model on the widely used MD17 benchmark dataset. Following established protocols~\cite{benchmark}, we selected four representative molecules known for their diverse chemical properties: aspirin, ethanol, naphthalene, and salicylic acid. To test the scalability of our ensemble method, we also utilized the large-scale MatPES dataset~\cite{matpes}. We partitioned this dataset into 90\% for training, 5\% for validation, and 5\% for testing. The model's ability to generalize was evaluated by performing structural relaxations on 1,000 structures from the WBM test set~\cite{wbm} and calculating their formation energies.

Finally, for a comprehensive validation of the learned force fields, we performed molecular dynamics (MD) simulations for the selected systems. These simulations provide critical metrics on the stability and practical accuracy of our model in reproducing dynamic properties.

\subsection{MLFFs}

In our work, we utilized a diverse and carefully selected set of MLFFs as base models, tailoring the specific ensemble to the unique chemical systems and predictive tasks of each dataset. For the experiments involving methane and methanol on a Cu(100) surface, we constructed an eight-model ensemble from four distinct frameworks. This included three models from DeepMD~\cite{DEEPMD}, each employing a unique descriptor (\texttt{se\_e2\_a}, \texttt{se\_e2\_r}, and \texttt{se\_e3}); three variations of SchNet~\cite{schütt2017schnet}, differentiated by their number of basis functions; a single model from the PaiNN framework~\cite{painn}; and one model from the Neuroevolution Potential (NEP)~\cite{nep} framework. When addressing the widely-used \textbf{MD17} benchmark, we aligned our methodology with previous research~\cite{benchmark} by adopting an identical set of MLFFs and parameters. This set comprised models from DeepMD, SchNet, DimeNet~\cite{dimenet}, ForceNet~\cite{forcenet}, and NequIP. Finally, for the materials-focused MatPES dataset, our base models were CHGNet~\cite{chgnet}, TensorNet~\cite{simeon2023tensornet}, M3GNet~\cite{m3gnet} and a reference model AlphaNet~\cite{yin2025alphanet}, which is not used in this method, all pre-trained on MatPES PBE data. These models were then evaluated on their ability to perform structural relaxations and calculate formation energies for a test set of 1,000 high-energy structures from the WBM database. This tailored approach ensured that the base models for each task were relevant and powerful, providing a robust foundation for our meta-learning framework.

\subsection{Metrics}
A comprehensive evaluation of a MLFF requires metrics that assess not only the point-wise accuracy of predictions, but also the physical realism of the resulting MD simulations. Drawing inspiration from the benchmark established by a previous study~\cite{benchmark}, which emphasizes the importance of stability and structural distributions, we employ a multi-faceted evaluation strategy. Our assessment is based on two categories of metrics: direct prediction accuracy and simulation fidelity.

To quantify the precision of point-wise prediction, we utilize two standard regression metrics: the Root Mean Square Error (RMSE) and the Mean Absolute Error (MAE). These are calculated as:
\begin{equation}
    \overline{\text{RMSE}} = \frac{1}{m}\sum_{j=1}^{m}\sqrt{\frac{1}{n_j}\sum_{i=1}^{n_j}(y_{ij} - \hat{y}_{ij})^2},
\end{equation}
\begin{equation}
    \overline{\text{MAE}} = \frac{1}{m}\sum_{j=1}^{m}\frac{1}{n_j}\sum_{i=1}^{n_j}\left| y_{ij} - \hat{y}_{ij} \right|.
\end{equation}
In these equations, $m$ is the number of systems, $n_j$ is the number of data points for system $j$, and $y_{ij}$ and $\hat{y}_{ij}$ are the reference and predicted values, respectively.

Beyond direct accuracy, a robust force field must generate stable trajectories and reproduce correct structural ensembles. To this end, we first evaluate the model's ability to replicate the distribution of interatomic distances $h(r)$, a low-dimensional descriptor of the 3D structure~\cite{hr}. For a given configuration $\mathbf{x}$, this distribution is calculated as
\begin{equation}
    h(r) = \frac{1}{N(N - 1)} \sum_{i=1}^{N} \sum_{\substack{j=1 \\ j \neq i}}^{N} \delta(r - \| \mathbf{x}_i - \mathbf{x}_j \|),
\end{equation}
where $N$ is the number of atoms and $\delta$ is the Dirac delta function. We then quantify the deviation between the model-predicted equilibrium distribution, $\langle \hat{h}(r) \rangle$, and the reference distribution, $\langle h(r) \rangle$, by calculating the integrated Mean Absolute Error:
\begin{equation} \label{eq:hr}
    \text{MAE}(h(r)) = \int_{0}^{\infty} \left| \langle h(r) \rangle - \langle \hat{h}(r) \rangle \right| dr.
\end{equation}

Finally, we assess the long-term simulation stability, a critical indicator of a model's practical utility, particularly for flexible molecules. While this property is often probed using observables like the radial distribution function (RDF) \cite{RDF1, RDF2}, we adopt a direct structural criterion by monitoring bond lengths. A simulation is deemed unstable if any bond length deviates from its equilibrium value by a threshold $\Delta$:
\begin{equation} \label{tab:stability}
    \max_{(i,j) \in B} \left| \| \mathbf{x}_i(T) - \mathbf{x}_j(T) \| - b_{i,j} \right| > \Delta,
\end{equation}
where $B$ is the set of all bonds, $\mathbf{x}_i(T)$ is the atomic position at time $T$, and $b_{i,j}$ is the equilibrium bond length. For all systems in this study, we set the threshold $\Delta = 0.5$ Å. Together, these metrics provide a holistic assessment of force field quality.

\section*{Conclusions}
In summary, we present EL-MLFFs, a systematic ensemble learning framework designed to address the challenges of model selection and reliability. By effectively leveraging the diverse predictive information from multiple base models, our framework synthesizes these inputs to yield highly reliable predictions, significantly enhancing both accuracy and generalization. With the flexibility of both a computationally efficient direct-fitting architecture and a physically rigorous conservative model, our approach successfully mitigates the limitations of individual force fields, ensuring robust dynamics across both molecular and condensed phase systems.

\section*{Code and data availability}
\label{sec:code}
Part of the codes and data are provided in \url{https://github.com/zmyybc/EL-MLFFs/tree/master}, the remaining part would be updated soon.

\bibliography{sample}

\end{document}